\documentclass[letterpaper, 10 pt, conference]{ieeeconf}  

\IEEEoverridecommandlockouts                              

\usepackage{graphics} 
\usepackage{epsfig} 
\usepackage{mathptmx} 
\usepackage{times} 
\usepackage{amsmath} 
\usepackage{amssymb}  
\usepackage[utf8]{inputenc}

\usepackage{color}
\usepackage{cite}
\usepackage{multicol}

\usepackage{subfigure}

\usepackage{algorithm}
\usepackage{algpseudocode}

\algnewcommand\algorithmicinput{\textbf{Input:}}
\algnewcommand\algorithmicoutput{\textbf{Output:}}
\algnewcommand\Input{\item[\algorithmicinput]}%
\algnewcommand\Output{\item[\algorithmicoutput]}%

\title{\LARGE \bf
Double Meta-Learning for Data Efficient Policy Optimization in Non-Stationary Environments
}

\author{Elahe Aghapour and Nora Ayanian
\thanks{Authors are with the Department of Computer Science, University of Southern California, USA 
        {\tt\small \{aghapour, ayanian\}@usc.edu}}%
}

\begin{document}

\maketitle
\thispagestyle{empty}
\pagestyle{empty}

\begin{abstract}
We are interested in learning models of non-stationary environments, which can be framed as a multi-task learning problem. Model-free reinforcement learning algorithms can achieve good asymptotic performance in multi-task learning at a cost of extensive sampling, due to their approach, which requires learning from scratch. While model-based approaches are among the most data efficient learning algorithms, they still struggle with complex tasks and model uncertainties. Meta-reinforcement learning addresses the efficiency and generalization challenges on multi task learning by quickly leveraging the meta-prior policy for a new task. In this paper, we propose a meta-reinforcement learning approach to learn the dynamic model of a non-stationary environment to be used for meta-policy optimization later. Due to the sample efficiency of model-based learning methods, we are able to simultaneously train both the meta-model of the non-stationary environment and the meta-policy until dynamic model convergence. Then, the meta-learned dynamic model of the environment will generate simulated data for meta-policy optimization. Our experiment demonstrates that our proposed method can meta-learn the policy in a non-stationary environment with the data efficiency of model-based learning approaches while achieving the high asymptotic performance of model-free meta-reinforcement learning.
\end{abstract}

\section{INTRODUCTION}

Many robotic control tasks can be captured as reinforcement learning (RL) problems, where the task objective function is being optimized by using collected data. RL approaches have been applied to a wide range of applications, from autonomous navigation \cite{AbbeelAutonomoushel, ZhangRLUAV2016} to manipulation \cite{KalakrishnanLfcpolicy2011, ShixiangDRLrobot2017}. 

Model-free RL algorithms are able to outperform human performance in many applications, but they are known to be data inefficient \cite{KamtheDataeffic2018, DulacrealWorld2019} and must be trained from scratch for a new task, which makes it impractical for many real world implementations.  
Moreover, RL methods demonstrate admirable performance in simulation \cite{SilverGameGo2016, LevineE2E2016, MnihAtari2013}, however, their performance may degrade or even fail in slightly different test environments \cite{BousmalisRobGrasp2018}; this is exacerbated in non-stationary environments, since the number of collected experiences, or data samples, is limited before changes occur.

For many robotic control tasks, collecting data is costly and the test environment is often different than the training environment due to non-stationary conditions, simulation to real world gap, and other perturbations. 
Hence, the robots' capability to quickly generalize to a new condition is critical. 
One technique to enable robots to use data efficiently under non-stationary conditions is meta-learning. 
Meta-learners are able to learn a meta-prior over model parameters of given tasks that can be quickly generalized to a new task using small amount of data \cite{schmidhuber2015learning, RL2Duan2016, VariBAD2019}.

In this paper, we build on gradient-based meta-learning methods \cite{MetaGrad2016, Hochreiter2001, Ravi2016, Nichol2018}. 
We apply model agnostic meta-learning (MAML) to meta-learn a policy and dynamic models. 
MAML, which was first proposed in \cite{FinnMAML2017}, is a gradient-based meta-learning algorithm with two connected phases: 1) meta-training learns a meta-prior model that can be quickly leveraged for new tasks, and 2) meta-testing adapts the model to the current new task. Many have taken inspiration from MAML \cite{Antoniou2018, Nichol2018meta, NagabandiDonline2018, LiFewShot2017, GrantHierarchical2018} due to its efficiency and flexibility.

The model-free (MF) meta-learning approaches tend to attain asymptotically optimal performance at the expense of data inefficiency \cite{FinnMAML2017, WangLearning2016, MishraASimple2017, RothfussPromp2018}. 
On the other hand, model-based (MB) meta-learning methods are data efficient, achieved by learning the dynamic model of environment instead of policy optimization \cite{ClaveraLearning2018, NagabandiDonline2018, NagabandiFineTuning2018}. However, learning an accurate dynamic model is often more complex than learning good policies. The performance of most model-based methods relies on accurate learned dynamic models. If the learned dynamic model is not accurate enough, it could lead to model bias. Then, the learned policy using the dynamic model could yield suboptimal performance or even failure. Recent works tried to reduce model bias by different approaches, e.g., by incorporating model uncertainty into action planning \cite{ZhouRobust1996, DeisenrothPILCO2011}, or model ensembles \cite{RajeswaranEpopt2016, LimrobustMDP2013}.

Finding an optimal policy in a non-stationary environment is challenging. 
A non-stationary environment can be treated as a sequence of stationary tasks where the model-free meta-policy optimization can be solved by continuous adaptation of meta-learning \cite{ShedivatConAdapt}. 
Meta-model-based RL algorithms enable sample-efficient learning \cite{ClaveraLearning2018, NagabandiDonline2018, NagabandiFineTuning2018}. 
The ability to adapt online in meta-learning alleviates the need to create a perfect model of a complex environment. 
However, due to the challenge of learning a sufficiently accurate dynamic model, 
model-based learning approaches struggle to accomplish complex tasks and achieve the super-human performance of model-free methods.

Taking advantage of data efficiency in model-based learning in combination with the asymptotic performance of model-free learning is an appealing idea for researchers. 
An alternative method is to accommodate the model-free policy optimization with learned dynamic model by generating trajectories to fine-tune the policy model initialization \cite{NagabandiFineTuning2018, GuQlearning2016}; however, these approaches still rely on large amounts of real-world data. In other works, model-based training and model-free training are decoupled and the model-free policy is trained by generated samples from learned dynamic model \cite{FeinbergMBvalue2018, KurutachModenEns2018}. However, these methods still rely on accurate learned models of environment. Another alternative is to use an ensemble of model-based reinforcement learners which are separately learning different dynamic models corresponding to different tasks \cite{ClaveraMBMP2018}. Then, an ensemble of learned models generate sample trajectories to train a meta-policy. Although this approach is more data efficient than training a model-free meta-policy, a sufficiently large amount of data from different tasks is required to train separate dynamic models corresponding to each task.

In this paper, we propose a double meta-reinforcement learning (DM-RL) approach to find a meta-policy in a non-stationary environment when data is costly. 
Our approach has two training phases. 
\textbf{Phase 1:} The meta-dynamic model and meta-policy are concurrently meta-trained by collecting samples from the environment while actions are selected by the learned meta-policy. When the meta-prior for the dynamic model converges, we move to phase 2. 
\textbf{Phase 2:} The meta-policy is trained using meta-data generated by the learned dynamic model. Then, 
the learned meta-policy can be quickly adapted to uncertainties and perturbations between the learned model and the current context in the environment at test time. 
DM-RL is significantly more data efficient than training a meta-policy \cite{FinnMAML2017} and training separate model-based RLs \cite{ClaveraMBMP2018}, while its performance closely matches the model-free meta-policy learning approaches. In addition, it often outperforms model-based approaches using MPC \cite{ClaveraLearning2018} on challenging tasks.

\section{PRELIMINARIES\label{sec:prelim}}
%
\subsection{Definitions}
We consider learning problems in discrete time Markov Decision Processes (MDP) described by a tuple $\{ S,~A,~\mathcal{P},~r,~ p_0,~ \gamma, ~ \}$, where $S \in \mathbb{R}^n$ is the state space,
$A \in \mathbb{R}^m$ is the action space, 
$\mathcal{P} : S \times A \Longrightarrow S$ is the state transition function, 
$r : S \times A \Longrightarrow \mathbb{R}$ is the reward function, $p_0$ is the initial state distribution, $\gamma$ is a discount factor, and $H$ is the length of time horizon. 
A trajectory of length $L$ is denoted by $\tau= \{s_0,a_0, \cdots, s_{L-1}, a_{L-1}, s_L \}$ where $s_k$ and $a_k$ denote state and action at time $k$, respectively. 
The cardinality of a set $G$ is denoted by $|G|$ and the p-norm of a vector $V$ is represented by $\|V \|_p$.

\subsection{Meta-Policy Learning}
The return is the discounted sum of the expected reward from a trajectory 
(i.e., $R(\tau) = \sum_{(s_t, a_t) \in \tau} {\gamma}^{\:t} r(s_t,a_t)$). 
The reinforcement learning goal is to find an optimal policy $\pi: S \Longrightarrow A$ that selects an optimal action in each state to maximize the expected return. The policy will be modeled by a neural network with inertial weights $\Phi$, and its weights can be found by minimizing the RL objective function:
\begin{equation}\label{eqn:RLobj}
\begin{split}
    L(\Phi, D) = & -\frac{1}{|D|} \sum_{t=0:|D|} \gamma^{\:t} \, r(s_t,a_t), \\
    s_0 \sim & p_0, \; s \sim p(s'|s,a), \; a_t \sim \pi_{\Phi} (a_t|s_t),
\end{split}
\end{equation}
where $D$ is training data and the loss function $L$ is a function of $D$ and $\Phi$. 

While deep reinforcement learning struggles with data inefficiency and limited generalization, 
meta-reinforcement learning is a data efficient approach which learns how to extend past learned policies to new tasks.
The meta-reinforcement learning algorithm will find a meta-prior parameter $\Phi$ which can be adapted to new tasks quickly. 
During meta-training, we have a task distribution $p(\mathcal{T})$ from which tasks $\mathcal{T}^i$ are sampled. Each task can be defined by a different MDP tuple $\mathcal{T}^i = \{S^i,~ A^i, ~p^i(s_{j+1}|s_j,a_j), ~r^i(s_j,a_j), ~p_0^i ,~ \gamma,~H \}$. 
Here, our focus is on learning in non-stationary environments; this is analogous to a distribution of tasks represented by MDPs 
where all tasks share the same state space $S^i = S$, action space $A^i = A$, reward function $r^i = r$, and initial state distribution $p_0^i = p_0$ while the transition function varies over different tasks (e.g. $\mathcal{T}^i = \{S, ~A, ~p^i(s_{j+1}|s_j,a_j), ~r(s_j,a_j), ~p_0,~ \gamma,~H \}$).
Our proposed solution is built upon the gradient-based meta-learning framework. We apply the model agnostic meta-learning (MAML) framework \cite{FinnMAML2017} where the training data $D^i$ from task $\mathcal{T}^i$ is split into meta-train data $D_{tr}^i$ and meta-test data $D_{ts}^i$. The MAML objective function is looking for $\Phi$ such that:
\begin{equation}\label{eqn:MAMLobj}
    \min_{\Phi} \sum_{\mathcal{T}^i \sim p(\mathcal{T})} L \big(\Phi - \alpha \nabla_{\Phi} L(\Phi, \, D_{tr}^i), \, D_{ts}^i \big)
\end{equation}
where $\alpha$ is learning rate hyper-parameters. An algorithm to solve \eqref{eqn:MAMLobj} is discussed in \cite{FinnMAML2017}.

\begin{figure*}[htp]
  \centering
  \subfigure[\label{fig:phase1}Phase 1]{\includegraphics[scale=0.58]{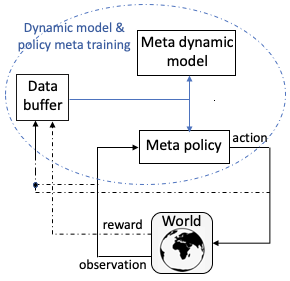}}\qquad\qquad\qquad
  \subfigure[\label{fig:phase2}Phase 2]{\includegraphics[scale=0.58]{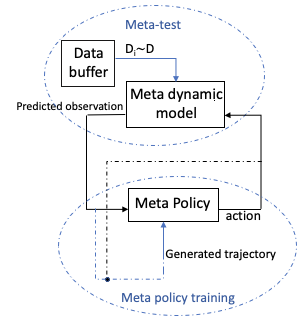}}
  \caption{Data flow diagram of double meta-learning algorithm. The blue and black lines represent lower and higher frequency of data, respectively. In phase 1, at each time instant, the meta-policy must choose the next action, interacting with environment (high frequency black lines) while meta-policy and meta-dynamic models are updated using batch of accumulated data (low frequency blue lines). In phase 2, at each time instant, meta-policy chooses the next action, interacting with meta-dynamic model (high frequency black lines). However, The meta-dynamic model updates once before generating a batch of data and meta-policy updates once using batch of simulated data (low frequency blue lines). }
\end{figure*}

\subsection{Learning The Prior Model By Meta-Training}
In model-based RL methods, we would like to learn the transition function $\mathcal{P}$ of the environment. Here, the dynamic model is represented by a Neural Network (NN) with internal weights $\theta$ that takes a pair $\{ s_t, a_t \}$ as input and predicts the state $s_{t+1}$. 
Given data $D$, the loss function to find internal weights $\theta$ is:
\begin{equation}\label{eqn:NNweight}
    L(\theta, D) = \frac{1}{|D|} \sum_{(s_t, a_t, s_{t+1}) \in D} {\| s_{t+1} - p_{\theta}(s_t,~ a_t)\|_2}^2.
\end{equation}

For a non-stationary environment, the meta-dynamic model is required to obtain $\theta_0$ that can be quickly leveraged to a new task with few gradient steps. 
Here, we rely on the model-based meta-reinforcement learning approach detailed in \cite{ClaveraLearning2018} to meta-learn $\theta$ such that:
\begin{equation}\label{eqn:RLobjMB}
\begin{split}
    \max_{\theta} \; &  \frac{1}{|D^i|} \sum_{\mathcal{T}^i \sim p(\mathcal{T})} \sum_{t=0:N} \gamma^t r(s_t,a_t) \\
    \textit{where} &\quad s_0 \sim p(s_0), s' \sim p_{\theta}(s,a).
\end{split}
\end{equation}
The proposed solution for \eqref{eqn:RLobjMB} iterates between two steps:
1) meta-learn the transition function model $P_\theta$ in \eqref{eqn:MAMLobj} using given data and MAML algorithm in \cite{ClaveraLearning2018}, and 
2) use model predictive control (MPC) and learned model $p_{\theta}$ to find  a sequence of actions with maximum return. The first action $a_t$ will be executed, and leads to state $s_{t+1}$. The tuple $\{s_t,a_t,s_{t+1} \}$ will be added to dataset $D^i$ to update the meta-prior model $P_\theta$.  

Solving MPC imposes high computational cost which leads to a sampling based MPC scheme:
$n_{candidate}$ random candidate action sequences of length $H$ are generated, and the sequence of actions with the highest predicted reward is selected. The reward of the action sequence is computed with the given reward function and learned dynamic model. 
The sampling-based MPC solution has sub-optimal performance and could even lead to failure, especially in high dimensional or continuous action spaces. 
Here, our main goal is to be as data efficient as model-based meta-RL with optimal performance as good as meta-policy algorithms.

\section{DOUBLE META-LEARNING FOR POLICY OPTIMIZATION \label{DoubleMetaL}} 

In this section, we propose a DM-RL approach to learn the meta-policy parameter $\Phi$ in a non-stationary environment. 
A non-stationary condition can be captured as a sequence of tasks where the problem of learning is seen as a few shot learning problem: we can collect limited samples from the environment before it changes. 
We assume that during training, a distribution of tasks $p(\mathcal{T})$ is given, from which tasks $\mathcal{T}^i$ are drawn. 
%
Since the on-policy methods can learn exploration strategies during meta-training, we collect data samples $D^i$ corresponding to task $\mathcal{T}^i$ using meta-RL policy $\pi_\Phi$. 
We seek meta-dynamic model parameter $\theta$ and meta-policy parameter $\Phi$ such that: 
\begin{equation}\label{eqn:RLobj2}
\begin{split}
    \max_{\theta, \, \Phi} \; &  \frac{1}{|D^i|} \sum_{\mathcal{T}^i \sim p(\mathcal{T})} \sum_{t=0:|D^i|} \gamma^t r(s_t,a_t) \\
    \textit{where} &\quad s_0 \sim p(s_0), \, s' \sim p_{\theta}(s,a), \, a \sim \pi_\Phi(a,s),
\end{split}
\end{equation}
where the reward function $r(s,a)$ is given. 
The optimization solution can be divided into two sub-problems:
\begin{itemize}
    \item \textbf{Policy fitting:} Applying meta-learning approaches to meta-train policy parameter $\Phi$. Here,  $\Phi$ is learned using the gradient based meta-learning approach described in \cite{FinnMAML2017}.
    
    \item \textbf{Dynamic model fitting:} Meta-training the dynamic model of the environment $\theta$ using data sample $D^i$s. Here, gradient based meta-learning is used as described in 
    \cite{ClaveraLearning2018}.
\end{itemize}
Note that these two sub-problems can be solved concurrently. 
The algorithm to solve optimization \eqref{eqn:RLobj2} is outlined in Algorithm~\ref{algorithm:algorithm1}, where the policy is meta-trained in lines 7 and 10 and the dynamic model is fitted by lines 8 and 11.    
We store data $D^i$ collected from the real-world in a buffer $D$ and repeat Algorithm~\ref{algorithm:algorithm1} until the meta-learned dynamic model converges. Figure~\ref{fig:phase1} demonstrates the algorithm described here. 

Moving into the next phase, the learned model $P_\theta$ will be used to generate simulated data until meta-policy parameter $\Phi$ satisfies the performance condition. 
The meta-learned dynamic model is capable of quickly adapting to new tasks at test time. 
The pool of samples $D_f$ is assumed as the true distribution of data. 
We uniformly draw data trajectories $D^i$s from the data buffer $D_f$, 
then selected $D^i$ will be used for dynamic model adaptation $P_{\theta^i}$ to task $\mathcal{T}^i$. Afterwards, the adapted dynamic model $P_{\theta^i}$ can generate simulated data to meta-train policy parameters $\Phi$. 
This algorithm is described in Fig.~\ref{fig:phase2} and Algorithm~\ref{algorithm:algorithm2}. 

As discussed in Achille et al., the early transient in training a neural network is critical in determining the final solution and its convergence~\cite{AchilleCritical2017}. Here, the meta-policy parameters $\Phi$ are initially meta-trained directly by true environment data. 
The generated data by $P_\theta$ is not exactly the same as true environment data due to model inaccuracy. However, meta-policy parameters $\Phi$ are capable of fast adaptation to new tasks and overcome small errors between the learned dynamic model and real data at test time. 

\begin{algorithm}
\begin{algorithmic}[1]
\Input Distribution over tasks $p(\mathcal{T})$
\Input Learning rates $\alpha$ and $\beta$
\State Randomly initialize the dynamic model $P_\theta$ and the policy $\pi_\Phi$
\State Initialize data buffer $D_f = \emptyset$
\While{not done}
    \State Sample batch of tasks $\mathcal{T}^i$ from $p(\mathcal{T})$
    \For{all T}
        \State Sample trajectories $D^i$ using policy $\pi_{\Phi}$ and $D_f = D_f \cup D^i$ split $D^i$ into $D^i_{tr}$ and $D^i_{ts}$
        \State Update $\Phi = \Phi - \beta \, \nabla_\Phi \, L(\Phi , \, D^i_{tr})$
        \State Update $\theta = \theta - \alpha \, \nabla_\theta \, L(\theta , \, D^i_{tr})$
    \EndFor
    \State $\Phi \longleftarrow  \Phi - \beta \, \nabla_\Phi \frac{1}{|\mathcal{T}|}\sum_{\mathcal{T}^i} L(\Phi , \, D^i_{ts})$
    \State $\theta \longleftarrow  \theta - \alpha \, \nabla_\theta \frac{1}{|\mathcal{T}|}\sum_{\mathcal{T}^i} L(\theta , \, D^i_{ts})$
\EndWhile
\Output Meta-learned dynamic model $P_\theta$ and meta-policy parameters $\Phi$, data buffer $D_f$
\end{algorithmic}
\caption{Phase 1: Meta-learning dynamic model and policy}
\label{algorithm:algorithm1}
\end{algorithm}

The standard approach to train a model-free RL using the data efficiency of model-based RL is to alternate between model learning and policy optimization. 
In learning the model, environment samples are used to fit the dynamic model then the learned model is used to search for policy improvement \cite{KumarOptCont2016, LevineLearningNNpol}. This setting can work well with low dimensional simple environments. 
However, it could be highly unstable in more challenging continuous control tasks since the policy tends to exploit the regions where insufficient data is available \cite{KurutachModenEns2018}. 
Here, to prevent this instability, the meta-policy is being used as a controller to explore and exploit the environment in phase 1 and both meta-policy and meta-dynamic models are being meta-learned  using the same environment data.

\section{EXPERIMENTS\label{sec:experiments}}
The main goal of this section is to study the following questions: 
\begin{enumerate}
  \item Does our proposed solution successfully train a meta-policy which can be quickly adapted at test time? How robust is our proposed method when it experiences a new task that is outside the distribution of the training tasks?

  \item How does our proposed algorithm perform in comparison with model-based RL MAML and model-free RL MAML with respect to sample efficiency and performance? How quickly does our solution adapt to a new task at test time in comparison with model-based RL MAML and Model-free RL MAML?
\end{enumerate}

\begin{algorithm}
\begin{algorithmic}[1]
\Input Meta-learned dynamic model $P_\theta$, learning rate $\alpha$
\Input Meta-policy parameters $\Phi$, learning rate $\beta$
\Input Data buffer $D_f$
\While{not done}
    \State Randomly choose batch of $D^i$s from $D_f$
        \For{all $D^i$}
            \While{not done}
            \State Use $D^i$ to update $\theta$ corresponding to $\mathcal{T}^i$ by: $$\theta^i \longleftarrow \theta - \alpha \, \nabla_\theta L(\theta , \, D^i)$$
            \EndWhile
            \State Sample batch of data $\bar{D}^i$ using dynamic model $P_{\theta^i}$ and policy $\pi_\Phi$
            \State Update $\Phi = \Phi - \beta \, \nabla_\Phi L(\Phi , \, \bar{D}^i)$
        \EndFor
    \State Update $\Phi \longleftarrow  \Phi - \beta \, \nabla_\Phi \frac{1}{|\mathcal{T}|}\sum_{\mathcal{T}^i} L(\Phi , \, \bar{D}^i)$
    \EndWhile
\Output Meta-policy parameters $\Phi$
\end{algorithmic}
\caption{Phase 2: Learning meta-policy from simulated data}
\label{algorithm:algorithm2}
\end{algorithm}

We evaluate the performance of DM-RL in comparison with the following methods:
\begin{itemize}
    \item \textbf{Model-free meta-reinforcement learning (MF-MAML-RL):} to evaluate data efficiency, we compare with the model-free model agnostic meta-learning (MAML) RL method, as described in \cite{FinnMAML2017}.
    \item \textbf{Model-based meta-reinforcement learning (MB-MAML-RL):} To study the computational cost and final performance, we implemented the model-based model agnostic meta-learning (MAML) RL method where model-based MAML RL uses MPC to choose its future action as described in \cite{ClaveraLearning2018}.
\end{itemize}

For the sake of consistency, the hyper-parameter settings for meta-policy learning and meta-dynamic model learning in DM-RL is chosen to be the same as MF-MAML-RL and MB-MAML-RL hyper-parameters, respectively. 
The MPC parameters and implementation is the same as the original paper \cite{ClaveraLearning2018}.

\subsection{Implementation Setup}
The main motivation of our work is to propose a learning-based controller for unmanned aerial vehicles, flying in highly dynamic environments. 
This requires an environment to reflect our real world problem. 
Our evaluation environment is derived from OpenAi Gym's LunarLander-v2 environment where the goal is to train the lunar lander to safely land on randomly generated specified surfaces on the moon~\cite{BrockmanOpenAIGym}. 
To adapt the LunarLander environment to be a satisfying option for meta-learning experiments, we add a wind generator function that generates wind with different speeds along the x-y axis. Different wind speed corresponds to a different state transition function, and subsequently a different task.
 

During training, a suite of tasks is defined by generating winds with different speeds along the x-y axis, drawn from a uniform distribution $U[-2,~2]$. The wind speed is assumed to be static throughout the duration of each rollout. For test time, we have two different scenarios: 1) Generating wind with constant speed drawn from the same uniform distribution during training but the lander has not seen it in its training data; and 2) Generating sinusoidal wind to study whether LunarLander can learn to extend its past experiences to a new situation which is different than its training distribution. 

In terms of hyper-parameter settings of our proposed algorithm, the dynamic model is trained on a four layer neural network with two hidden layers of size 64 and 32. 
The activation function is Rectified Linear Unit (ReLU) and both ReLU and softmax are used in the output layer.  
The policy model is also represented by a four layer neural network with two hidden layers of size 64. The activation function is ReLU and categorical for the output layer. 
We applied TRPO as the meta-optimizer \cite{SchulmanTRPO2015}. 
We utilize finite differences to compute the Hessian-vector products for TRPO in order to avoid third derivatives. 
Table \ref{Tab:hyperparam} shows the value of the hyper-parameters.

\begin{table}[th]
\caption{Value of hyper-parameters} \label{Tab:hyperparam}
{\begin{tabular}{ll} \hline
\textbf{Hyper-parameters} $\qquad \qquad \qquad \qquad \qquad$ &\textbf{Value} \\
\hline
Meta-batch size & 10 \\
$n_{candidate}$ (MPC)& 1000 \\
Horizon length H (MPC) &10 \\
Maximum rollout length  & 150 \\
Number of iterations & 200 \\
Monte Carlo trials & 10 \\
\hline
\end{tabular}}
\end{table}

\subsection{Results}

\begin{figure}[tp]
    \centering
    \includegraphics[width=0.5\textwidth]{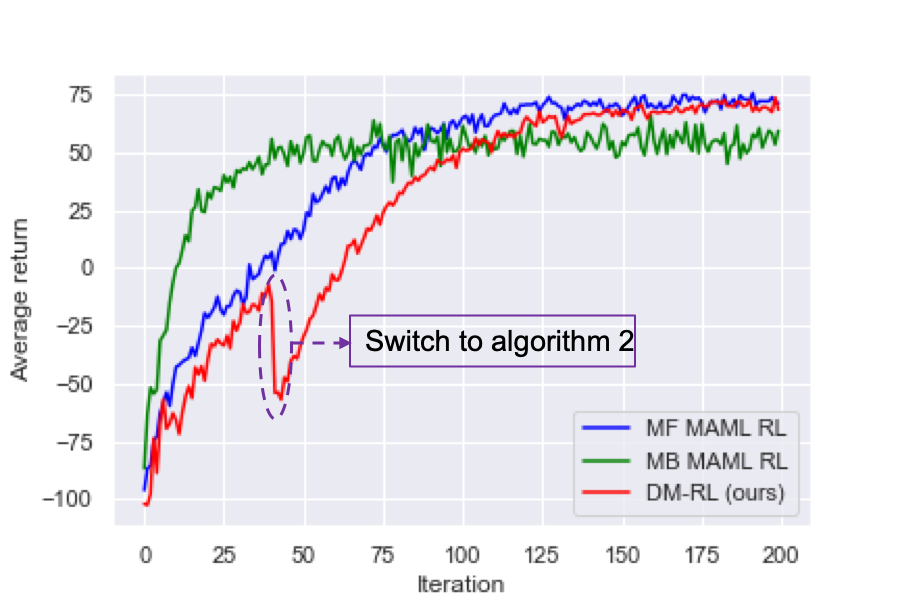}
    \caption{\label{fig:TrainData} Training curve of DM-RL in comparison with model-free MAML RL and model-based MAML RL averaged over 10 trials. DM-RL switches to algorithm 2 at iteration$=40$ which means it does not use true environment data anymore. DM-RL is able to closely match the asymptotic performance of MF methods with fewer true environment samples.}
\end{figure}

Figure \ref{fig:TrainData} shows the average returns of meta-batches during meta-training. Each iteration uses one meta-batch of data. 
The red curve is meta-policy return in DM-RL. The blue and green curves represent MF-MAML-RL and MB-MAML-RL, respectively. 
Both MF MAML and MB MAML are only using true environment data and their return increases over time until convergence. The MF approach  achieves good asymptotic performance but at the cost of extensive data, converging around $iteration = 140$, while the MB approach exhibits data efficient learning and converges around $iteration=40$. However, due to model uncertainty and suboptimality in the sample-based approximation of MPC, it struggles to perform well on accomplishing tasks that require robust planning.
 
For the first 40 iterations, the meta-policy of DM-RL is being trained by Algorithm~\ref{algorithm:algorithm1}, which uses true environment data, and its return increases over time. 
At iteration 40, once the meta-dynamic model converges, we switch to phase 2 
(Algorithm~\ref{algorithm:algorithm2}), where the meta-learned dynamic model will generate simulated data. 
This switching from environment data to simulated data, combined with learned model inaccuracies, causes a drop at $iteration = 40$.  
Then, using generated data, the meta-policy return again increases to convergence. Although DM-RL needs more training iterations for convergence, only the first 40 iterations use true environment data. The performance of all three approaches is summarized in Table~\ref{Tab:trainPer}. 

\begin{table}[t]
\caption{Performance comparison at training.\label{Tab:trainPer}} 
{\begin{tabular}{llll} \hline
 & MF-MAML-RL & MB-MAML-RL & DM-RL  \\
\hline
Return mean & \textbf{71.59} & 55.08 &\textbf{69.67}  \\
after convergence & & & \\
\hline
Required batches & 1400 & 400 & 1600\\
 to converge & & & \\
\hline
Required batches  &1400& 400& \textbf{400} \\
from env to converge & & & \\
\hline
\end{tabular}}
\end{table}

\begin{figure}[tp]
    \centering
    \includegraphics[width=0.5\textwidth]{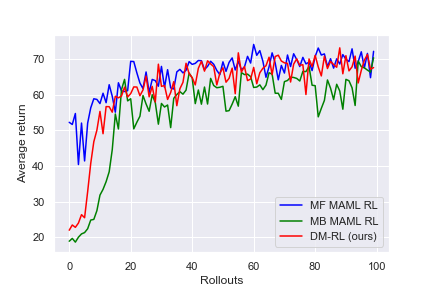}
    \caption{\label{fig:TestDataStat}\textbf{First scenario (static environment):} The comparison of returns at test time. Both the model-free MAML RL and DM-RL achieve the asymptotic performance.}
\end{figure}

\begin{table}[t]
\caption{Performance comparison at test time for first scenario \\ (static environment)} \label{Tab:TestStatPer}
{\begin{tabular}{llll} \hline
 & MF-MAML-RL & MB-MAML-RL & DM-RL \\
\hline
Return mean & 69.08 & 60.36 & 67.39  \\
after convergence & & & \\
\hline
Required batches & 20 & 25 & 25\\
 to converge & & & \\
\hline
Required batches  &20& 25& 25 \\
from env to converge & & & \\
\hline
\end{tabular}}
\end{table}

At test time, we evaluate the capability of the meta-learned models to leverage to a new task. The \textbf{first scenario} is in a new static environment. 
We applied the meta-learned approaches in an environment where new constant wind was selected from the range $[-2,2]$, but none of the learning models had seen it during training. 
Figure \ref{fig:TestDataStat} shows the performance of each algorithm. The MF-MAML-RL uses 20 rollouts to quickly adapt to the new task. DM-RL starts with lower returns due to dynamic model uncertainty in the second phase of training, but achieves the MF performance, using 25 rollouts. The MB-MAML-RL starts similarly to our method due to learned dynamic model uncertainty, but its performance is not as robust as the other methods due to the approximation in sample based MPC implementation. Table \ref{Tab:TestStatPer} summarizes the comparison. Note that all three methods are only using environment samples at test time. Thus, last two rows of Table \ref{Tab:TestStatPer} and \ref{Tab:TestSinPer} have the same values. 

\begin{figure}[tp]
    \centering
    \includegraphics[width=0.5\textwidth]{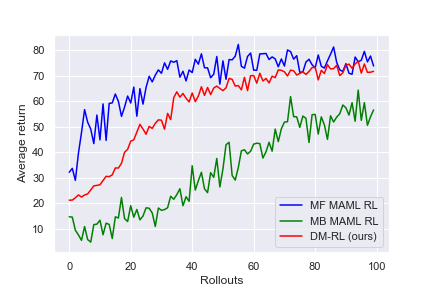}
    \caption{\label{fig:TestDataSin}\textbf{Second scenario (dynamic environment):} DM-RL outperforms model-based MAML RL when new task is dynamic and need online adaptation. Our method converges to model-free MAML RL within 70 rollouts.}
\end{figure}

The constant velocity assumption on generated wind is not realistic compared to real-world conditions where wind speed can change over time. To approach real-world conditions, in the \textbf{second scenario} the environment is dynamic. 
The generated wind speed $v$ is sinusoidal along the x-y axis:
$$v(t_k)= \begin{bmatrix} A_x \, \sin(\omega_x \, t_k) \\ A_y \, \sin(\omega_y \, t_k) \end{bmatrix}$$
with amplitude $A=2$ and frequency $\omega=0.01Hz$. The generated wind  remains in $[-2,2]$ but changes over time. 
The new given environment is dynamically different than the meta-training data. However, the meta-learned models must be capable of extending their past experiences to a new task quickly. Figure~\ref{fig:TestDataSin} shows the performance of the different approaches. MF-MAML-RL return increases over time and converges at $iteration=40$. DM-RL return starts lower than the model-free approach due to the learned model dynamic inaccuracy but achieves the asymptotic performance of MF-MAML-RL using 70 rollouts. Although DM-RL uses 30 more batches at test time to adapt to a new task, it uses only 400 batches from the environment during training while MF-MAML-RL used 1400 batches to be trained (i.e., more than three times the meta-training data). The MB-MAML-RL performance is not as good as the other methods since the environment is dynamic and it requires learning a dynamic model accurate enough to work with MPC. 
Performance values are summarized in Table~\ref{Tab:TestSinPer}.


\begin{table}[t]
\caption{Performance comparison at test time for second scenario (dynamic environment)} \label{Tab:TestSinPer}
{\begin{tabular}{llll} \hline
 & MF-MAML-RL & MB-MAML-RL & DM-RL \\
\hline
Return mean & 71.59 & 55.08 &69.67  \\
after convergence & & & \\
\hline
Required batches & 40 & 70 & 70\\
 to converge & & & \\
\hline
Required batches  &40& 70 & 70 \\
from env to converge & & & \\
\hline
\end{tabular}}
\end{table}

Our results  demonstrate that:
\begin{enumerate}
    \item DM-RL can successfully train a meta-policy that is quickly generalized to a new task at test time even when the new task was changing over time in the test environment (see Figures \ref{fig:TestDataStat} and \ref{fig:TestDataSin}); and
    \item DM-RL can be trained as data efficiently as model-based meta-RL  (see Figure \ref{fig:TrainData}) while achieving the  performance of model-free meta-RL at the cost of a few more trajectories from the test environment (see Figures \ref{fig:TestDataStat} and \ref{fig:TestDataSin}). 
    Since DM-RL trains the meta-policy with true environment samples at phase 1 and meta-data generated by the meta-dynamic model at phase 2, a few more trajectories at test time are required due to the difference between the data generated by meta-dynamic model and true environment data. However, it is still significantly more data efficient than MF meta-RL.  
\end{enumerate}


\section{CONCLUSIONS}
In this paper, we presented a simple and generally applicable approach to efficiently meta-learn a policy in a non-stationary environment. 
We combined the advantage of data efficiency in model-based learning with asymptotic performance of model-free approaches in the context of double meta-learning where initially the meta-dynamic model of the environment and meta-policy are concurrently being trained using environment data until the meta-dynamic model converges. Then, the meta-dynamic model is used to generate data to meta-train the policy. 

Due to applying meta-learning to optimize the policy in a dynamic environment, the meta-policy at test time is robust to environment changes and can be quickly adapted to new tasks. 
Our experimental results show that our proposed approach achieves the asymptotic performance of model-free meta-reinforcement learning approach with considerably smaller sets of meta-training data. 
We also evaluate our method against model-based meta-reinforcement learning while using the same amount of meta-training data; our method demonstrates better performance than the model-based method at test time. 
An exciting future direction is to apply our approach for real world system implementation. 
Our proposed approach relies on on-policy data, limiting its sample efficiency. Applying off-policy meta-reinforcement learning to our approach would also be of interest.


\end{document}